\documentclass[
]{ceurart}

\usepackage{listings}
\begin{document}

\copyrightyear{2024}
\copyrightclause{Copyright for this paper by its authors.
  Use permitted under Creative Commons License Attribution 4.0
  International (CC BY 4.0).}

\conference{ISWC: Special Session on LLMs}

\title{Information for Conversation Generation: Proposals Utilising Knowledge Graphs }
\tnotemark[1]
\tnotetext[1]{An extended version of this paper is available in arXiv.}

\author[1]{Alex Clay}[%
orcid= 0009-0004-2237-7412,
email=alex.clay@city.ac.uk,
]
\author[1]{Ernesto Jiménez-Ruiz}[%
orcid= 0000-0002-9083-4599,
email=ernesto.jimenez-ruiz@city.ac.uk,
]
\address[1]{City St George's, University of London,
Northampton Square, London, EC1V 0HB, United Kingdom}




\begin{abstract}
LLMs are frequently used tools for conversational generation. Without additional information LLMs can generate lower quality responses due to lacking relevant content and hallucinations, as well as the perception of  poor emotional capability, and an inability to maintain a consistent character. Knowledge graphs are commonly used forms of external knowledge and may provide solutions to these challenges. This paper introduces three proposals, utilizing knowledge graphs to enhance LLM generation. Firstly, dynamic knowledge graph embeddings and recommendation could allow for the integration of new information and the selection of relevant knowledge for response generation. Secondly, storing entities with emotional values as additional features may provide knowledge that is better emotionally aligned with the user input. Thirdly, integrating character information through narrative bubbles would maintain character consistency, as well as introducing a structure that would readily incorporate new information. 
\end{abstract}


\maketitle
\section{Introduction}
Large Language Models (LLMs) have quickly become the standard for the generation of conversational AI responses, due in part to the popularity of ChatGPT. However, without augmentation, LLMs are prone to a number of pitfalls, namely that of responses lacking valuable content~\cite{parthasarathi-pineau-2018-extending} and hallucinations~\cite{huang2023survey}, as well as lack of emotional capability~\cite{Chen}, and inconsistent character~\cite{xiao2024farllmsbelievableai}. External knowledge such as knowledge graphs (KG) can be used to improve the first two instances through supplying additional information, and may be key to resolving the latter two through retrieval augmented generation.

Responses lacking content and hallucinations are often due to incomplete information, as LLMs use information stored within their parameters from when they were trained~\cite{lewis-2020}. These challenges would still persist when utilizing static knowledge base, as out of date knowledge is known to contribute to hallucinations~\cite{huang2023survey}. Knowledge graphs are able to dynamically add information through the addition of new triples, but require embedding to predict new relationships between existing entities~\cite{10.5555/2999792.2999923}. As such, the adaptation of a knowledge graph embedding (KGE) to readily incorporate new information without requiring retraining would utilize the benefits of both KGs and KGEs, and better address content poor responses and hallucinations. The selection of the information is as crucial as the information itself. Therefore, a recommender could utilise the information stored in a KG to determine what would be most relevant to a user input.

Supplying the emotion through auxiliary information to the LLM at time of generation could provide a means for emotional integration. By selecting information that is more emotionally aligned with the user input through comparing emotional scores stored as features in the KG, it may be possible to generate a response that is more emotionally appropriate and increases the users' perception of the agent's emotional capability.

Additions of character to LLMs often utilize user supplied background information which guides the generation of character responses.Grounding an LLM with a knowledge base specifically for providing consistent character information could improve on these shallow character representations. We propose a structure to extract character information from novels, delineating between utterances, character facts, and providing summaries to act as character memory. This information would then be stored as entities within a KG and organized in a narrative bubble structure to maintain information about what was learned when and improve the integration of personal experience.

These proposals provide means to improve the user experience with conversational AIs. The active integration of new information would address frustration that users may encounter if the conversational AI displays a lack of knowledge~\cite{10.1007/978-3-031-35894-4_3} by allowing for knowledge base updates. Human-like AI is more likely to be accepted the user perceives it as having empathy~\cite{pelau-2021} or social emotion, which influences user trust and subsequently acceptance~\cite{zhang-2021}. By increasing the emotional alignment of responses and heightening the perception of emotional capability, the user may be more accepting of the AI. Consistency of character through storing character information could additionally improve user experience by potentially preventing instances where the LLM might hallucinate details or change information within the same interaction.

This paper will introduce proposals to utilise KGs to support the integration of new information, emotion, and character based personal experience into LLM generated responses. 

\section{Proposals for Information for Conversation Generation}

\subsection{New Information Integration and Recommendation}

Active integration of new information in knowledge bases is crucial for maintaining relevant and up-to-date information, and could reduce hallucinations on the part of the LLM~\cite{huang2023survey}. While new information can readily be added to knowledge graphs as they do not require training, the same is not true of KGEs, which require training to create embeddings. As language models can neither easily integrate new information or modify their knowledge~\cite{10.5555/3495724.3496517}, it is a crucial facet of the external knowledge base. Therefore, a Dynamic KGE (DKGE) would be ideal, as DKGEs are KGEs which are able to add new data into the embeddings without requiring retraining of the entire graph~\cite{cui2023lifelongembeddinglearningtransfer}, and could provide means to leverage the additional value of KGEs without the knowledge base being static. This reduces instances where the LLM lacks the necessary information to generate a correct response and causes frustration to the user~\cite{10.1007/978-3-031-35894-4_3}.

In order to utilize the information stored within the DKGE, a recommender could take the user input and supply relevant information to the LLM at the time of generation.  Knowledge Graph Embeddings (KGEs) have been utilised for recommendation in a number of approaches~\cite{Ai_2018, 10058002,GUO2020263}, which highlights the feasibility of integrating recommendation with a DKGE.

In recommendation, typically there are users and products with the intention of using available information to recommend the best product to the target user~\cite{frej2024graphreasoningexplainablecold}. When a user is previously unknown or otherwise lacks interaction and preference information, it is regarded as a cold start problem~\cite{10.1145/3397271.3401426}. To utilise this approach for information recommendation, the user input would take the role of an unknown user, for which supporting factual information would be provided instead of a product recommendation. This would be achieved through the use of a link prediction task where utterances and information are stored as different entity types in a heterogeneous knowledge graph, and a relation is predicted between the user input as an utterance type entity to a piece of information.

For example, if the user were to initiate a conversation about dinosaurs they might make a statement such as ``There is evidence the T. Rex may have been as intelligent as a crocodile.". This statement would then be handled as an entity \texttt{(kg:utterance1, rdf:type, kg:Utterance)
(kg:utterance1, kg:text, "There is evidence the T. Rex may have been as intelligent as a crocodile.")} for which a tail would be predicted for the relation, \texttt{relevant\textunderscore to} for a piece of background information. This would ideally return entities like \texttt{(kg:T. Rex, rdfs:subClassOf, kg:CarnivorousDinosaur)} which would be given to the LLM as \texttt{A T. Rex is a carnivorous dinosaur} for incorporation into an information rich response.

In order to keep the information up to date, the utterance would be added as shown before, as well as the entities \texttt{T. Rex} and \texttt{crocodile} (if not already present) as follows \texttt{(kg:T. Rex, rdfs:subClassOf, kg:Dinosaur)} and \texttt{(kg:Crocodile, rdfs:subClassOf, kg:Reptile)}. The best relations found during the initial prediction for recommendation would be used to integrate the new entities. The embedding for the new entity would then be an average of the embeddings of the related entities. Such as the utterance entity\texttt{(kg:utterance1, kg:text, "There is evidence the T. Rex may have been as intelligent as a crocodile.")} starting with an embedding derived from that of \texttt{T. Rex} and \texttt{crocodile} if they were already present in the knowledge base.

\subsection{Emotional Features for Entities}
The display of emotion and emotional understanding are crucial for the success of  conversational AI due to their necessity in human conversation~\citep{rashkin-etal-2019-towards, 10.5555/3504035.3504125, liu2022empathetic}. Such cues are often expected by the user~\citep{Chen} and can increase the perception of friendliness and intelligence of an AI~\citep{ijcai2018p618}.

A potential means of integrating emotion into generated text is through influencing the supplemental information given to the LLM at the time of generation to be more emotionally in line with that of the user. In taking an otherwise purely factual knowledge base and adding emotional values as features to the entities prior to training, and to the user input at runtime. Supplying more specificity than emotional labels, Valence, Arousal, and Dominance (VAD) scores~\cite{RUSSELL1977273} are commonly used to denote emotions as a point in space and could provide rich feature information. A recommendation structure based off of this may yield better supplementary data. 

For instance, compare the statements ``Rosalyne is picky" and ``Rosalyne is meticulous". A human can recognise that while the meaning is very similar,  meticulous has a positive connotation and picky does not, however an LLM would not necessarily have a means of differentiating such words without emotional delineation. As another example, in a situation without the emotional values, a statement of ``The weather is lovely today" could theoretically return sunny and rainy with equal likelihood. However, with the introduction of an emotion value if the input above was assigned a value closer to that of pleasant, and with the expectation that sunny would have a closer value to pleasant than rainy, the recommended information would be more pertinent to the conversation. 

The introduction of the emotional score then not only increases the likelihood of the returned information being more emotionally similar to the input, but also supplies an additional feature to the input entity which may improve the cold start recommendation. 



\subsection{Character Dimension Through Experience From Novels }

Companies like Character.AI in industry supply a means by which to add a character atop an LLM ~\cite{wang-etal-2024-characteristic} through character information supplied by the user and likely achieved through a form of prompt engineering. These approaches can lead to generic responses that do not integrate relevant character information~\cite{wang-etal-2024-characteristic}. Moreover it is unlikely that the character information prompt is updated during interaction, instead remaining static and relying on further details to be generated in response to user input. This can cause inconsistency and might lead a lengthy conversation to become nonsensical.



A potential solution is to introduce character information from a structured storage to the LLMs generation of the conversational agent's responses. Such a method could allow for the incorporation of more character information and responses potentially more similar to that of the character. The use of KGs in such a situation would reduce the number of needed tokens for an initial prompt, as the character would not require an initial explicit definition, but rather incorporate relevant character information as needed throughout the interaction. KGs would also provide consistent information, rather then relying on generation to  supply details adhoc which may not be retained between responses. Additionally, providing supporting information from an auxiliary knowledge base allows for a larger amount of character information to be stored and integrated into conversation. A particular attribute of KGs that could further improve the quality of responses is the latent information in relations that KGs are able to utilise which could allow for richer integration of character information with the rest of the knowledge base.


Novels lend the opportunity to gain abundant character information as well as the chance to evaluate the approach against existing expectations for the character. In research, some similar investigations have been made, with~\cite{Bogatu2015ConversationalAT} deriving character information from biographies and~\cite{zhang-etal-2018-personalizing} which utilised a set of five sentences to provide the basis of the agents characterisation. \cite{shao-etal-2023-character} investigated a similar concept in supplying personal experience through constructing scenes with which to fine-tune an LLM. However, it stands that by structuring the stored character information like memory the conversational integration of this information may seem more natural. Additionally~\cite{10.5555/3237383.3237945} found that in conversation it may be necessary to indicate recall explicitly, therefore delineating between character facts and character memory may be crucial in supplying this information to an LLM to generate dialogue.

\begin{figure}
    \centering
    \includegraphics[width=0.7\textwidth]{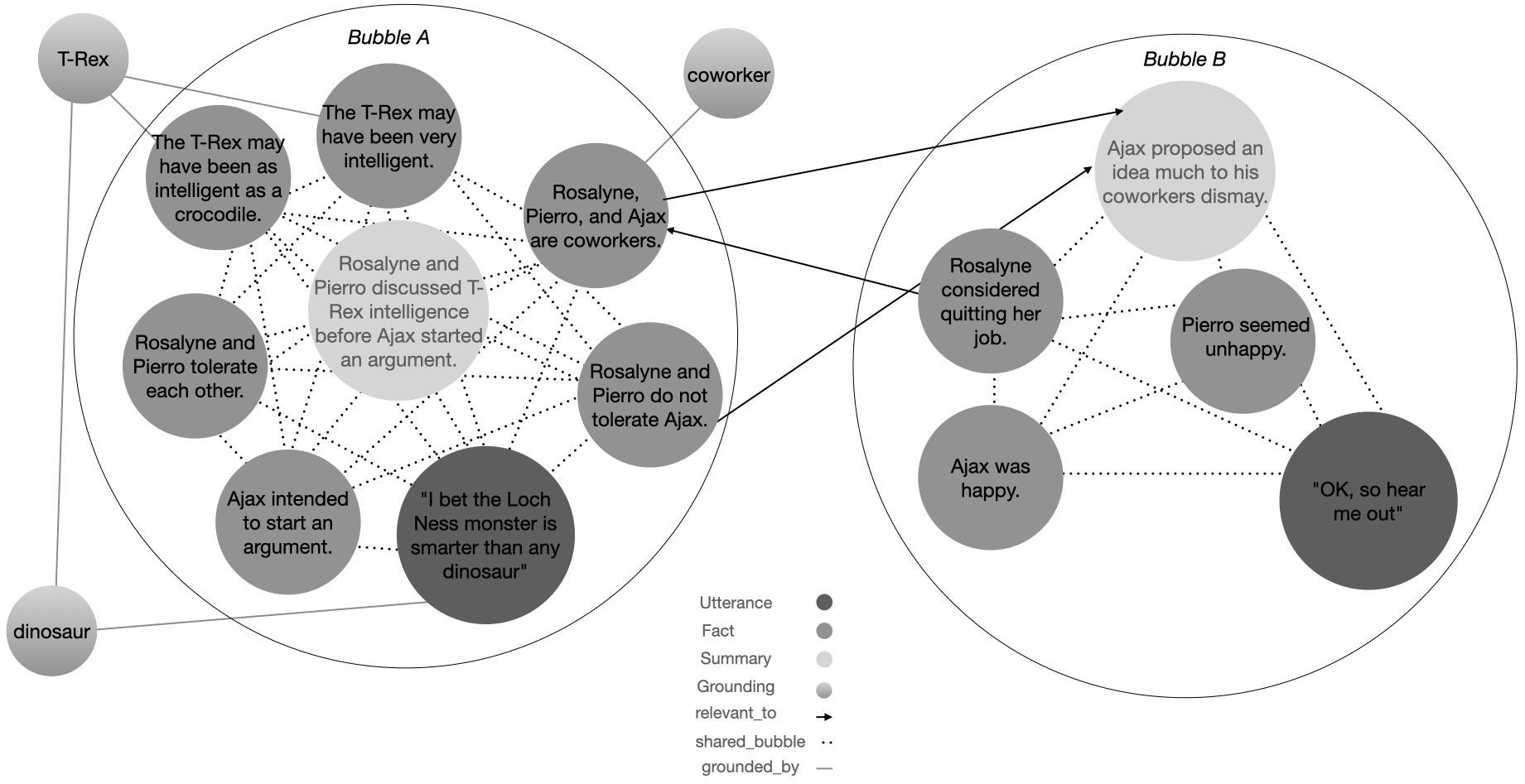}
    \caption{An example of two bubbles for the character Ajax. Each bubble only has one summary entity.}
    \label{fig}
\end{figure}

A novel could be structured into narrative `bubbles' within a KG for a particular character, wherein everything encountered during a particular spatial or temporal bound would be stored together, much like human personal experience~\cite{ezzyat-2011} which is known as  episodic memory and handles the ‘what’, ‘when’, and ‘where’ of experiences~\cite{tulving1972episodic, tulving-2002}. For example, the following text has been formatted as it might appear in the KG in Figure \ref{fig} for the character, Ajax. The two bubbles are split on the temporal delineation of ``the next day".
\begin{quote}
    ``The T-Rex may have actually been very intelligent," Pierro stated confidently.

    ``I heard otherwise," replied Rosalyne. ``They may have been as intelligent as a crocodile."

    ``I bet the Loch Ness monster is smarter than any dinosaur," Ajax interjected immediately starting an argument, exactly as he had intended. While Rosalyne and Pierro generally tolerated one another as much as one would a coworker, Ajax's shenanigans meant he did not receive the same courtesy.

    The next day was much the same.

    ``OK, so hear me out," Ajax began jovially. Pierro looked despondent. Rosalyne was actively considering quitting her job.
\end{quote}
The information within the bounds of the bubble would be separated into entities of the type utterances, facts, and an overall summary. The separation of this information is crucial as it can provide cues to the LLM about how to integrate the information appropriately, such as whether to indicate recall in a response.

Within the bubble each entity would have a relation to every other member entity regardless of type. This relation would be of a type \texttt{shared\textunderscore bubble} and act as the unifying factor of the bubble. Consider the first sentence of the third line of the example text. From the view of a KG constructed for the character Ajax, the first sentence would become two entities \texttt{(kg:utteranceA, rdf:type, kg:Utterance)(kg:utteranceA, kg:text, "I bet the Loch Ness monster is smarter than any dinosaur")} and \texttt{(kg:factA, rdf:type, kg:Fact)(kg:factA, kg:text, Ajax intended to start an argument")} which would form the triple \texttt{(kg:utteranceA, kg:shared\textunderscore bubble, kg:factA)} which indicates the entities shared presence within a bubble. 

Entity utterance1 would also be a member of the triple \texttt{(kg:utterance1, kg:grounded\textunderscore by, kg:Dinosaur)} relating to the grounding entity \texttt{dinosaur}, as shown in Figure \ref{fig}, which grounds it to information beyond the bubble relation and aids in informing its embedding through the relation \texttt{grounded\textunderscore by}.

In order to determine what relations should exist between bubbles, a comparison of the summary entities would be made and given enough relevance, the other entities within those bubbles would then be compared for a potential relation of the type \texttt{relevant\textunderscore to} which would not necessarily be reciprocal. For instance, the entity of type fact, \texttt{Rosalyne, Pierro, and Ajax are coworkers} forms a triple with the summary entity\texttt{ Ajax proposed an idea much to his coworkers dismay} with the relation of \texttt{relevant\textunderscore to} as the fact that they are coworkers provides useful supporting information.

For recommendation during a conversation, first a preliminary response would be generated for the user input. For example, if the user were to state ``what do you think about dinosaurs?" an LLM might generate ``dinosaurs are cool" as a response, which may not reflect character accurate information. This initial response would then be treated as an unseen utterance entity \texttt{"dinosaurs are cool"} for which link prediction for the relation \texttt{shared\textunderscore bubble} would be conducted, in this instance leading to a predicted link with the utterance entity \texttt{"I bet the Loch Ness monster is smarter than any dinosaur"} due to the shared presence of \texttt{dinosaur}. The contents of the \texttt{A} bubble in Figure \ref{fig} would then be returned as information for the LLM to use in generating a new character based response, possibly something like ``the Loch Ness monster is cooler than dinosaurs" replacing the initially generated response.




Another potential advantage of this approach would be the ability to dynamically add new bubbles. Following an embedding on the initializing information, a new bubble would first have relations formed with existing entities, the embeddings for which would then be aggregated to inform the embedding for the new entity. In a situation where an entity in a new bubble lacks any reference points from existing entities, its value would take an average from the other entities in the bubble. The summary entity of a bubble would always take its embedding from an aggregation of the entities in the bubble as it is the representation of the overall content. In order to reduce the need for retraining, existing entities would only update their embeddings after adding a certain threshold of new relations. If the bubble as a whole has had a number of updates to the member entities, the contents of the bubble would be reembedded with updated values as if it were a new bubble.

While integrating new information alone has been found to better support LLM generation, maintaining a structure focused on highlighting the relations between information learned and used together could support a more narratively focused information suggestion which could potentially more create more natural sounding dialogue.

\section{Conclusion}
LLMs provide a powerful tool for generating conversational responses, however their responses lack valuable content and hallucinations, as well as lack of emotional capability, and inconsistent character. This paper has introduced proposals for solving these challenges. Firstly, through utilizing dynamic knowledge graph embeddings with recommendation to provide up-to-date and relevant information to the LLM at time of generation to improve the content of responses and reduce likelihood of hallucinations. Additionally, using emotional values as features for entities to improve alignment with user input to increase the perception of emotional capability. Finally, the concept of storing character information in narrative bubbles was introduced, which could be updated without requiring retraining, to provide means for the portrayal of richer characters in LLM based conversation generation.

\bibliography{ref}

\end{document}